\documentclass[9pt,conference]{IEEEtran}
\usepackage{amsmath}
\usepackage{amsfonts}
\usepackage{amsthm}
\usepackage{graphicx}

\usepackage{url}
\usepackage{algorithm}
\usepackage{algpseudocode}
\usepackage{xcolor}
\usepackage{lipsum}
\usepackage{multirow}
\usepackage{booktabs}

\usepackage{float}
\usepackage{subcaption}

\newcommand{\R}{\mathbb{R}_{+}}

\DeclareMathOperator*{\argmin}{argmin}

\begin{document}

\title{Stratified Non-Negative Tensor Factorization}

\author{
    \IEEEauthorblockN{
    Alexander  Sietsema\IEEEauthorrefmark{1}\textsuperscript{\textsection}, Zerrin Vural\IEEEauthorrefmark{1}\textsuperscript{\textsection},
    James Chapman\IEEEauthorrefmark{1}, 
    Yotam Yaniv\IEEEauthorrefmark{2}, Deanna Needell\IEEEauthorrefmark{1}}
    \IEEEauthorblockA{\IEEEauthorrefmark{1}Department of Mathematics, University of California, Los Angeles}
    \IEEEauthorblockA{\IEEEauthorrefmark{2}Lawrence Berkeley National Laboratory}
}

\maketitle
\begingroup\renewcommand\thefootnote{\textsection}
\footnotetext{Equal contribution}
\endgroup

\begin{abstract}
    Non-negative matrix factorization (NMF) and non-negative tensor factorization (NTF) decompose non-negative high-dimensional data into non-negative low-rank components. NMF and NTF methods are popular for their intrinsic interpretability and effectiveness on large-scale data. Recent work developed Stratified-NMF, which applies NMF to regimes where data may come from different sources (strata) with different underlying distributions, and seeks to recover both strata-dependent information and global topics shared across strata. Applying Stratified-NMF to multi-modal data requires flattening across modes, and therefore loses geometric structure contained implicitly within the tensor. To address this problem, we extend Stratified-NMF to the tensor setting by developing a multiplicative update rule and demonstrating the method on text and image data. We find that Stratified-NTF can identify interpretable topics with lower memory requirements than Stratified-NMF. We also introduce a regularized version of the method and demonstrate its effects on image data.\end{abstract}

\section{Introduction}
Unsupervised methods for understanding large, multi-modal datasets have become increasingly important as data complexity has continued to grow. Non-negative matrix factorization (NMF) and non-negative tensor factorization (NTF) for higher-mode data are popular methods for factorizing large-scale data into interpretable, low-rank components \cite{shashua2005non, lee2000algorithms, lee1999learning}. Given a non-negative data matrix $A \in \R^{m \times n}$, NMF decomposes the data into non-negative factor matrices $W \in \R^{m \times r}$ and $H \in \R^{r \times n}$ that minimize the objective
\begin{equation}
\argmin_{W,H} \|A - WH\|^2_F.
\label{eqn:ntfloss}
\end{equation}

In this work, the data matrix is organized with rows corresponding to samples and columns corresponding to variables. The topics matrix $H$ gives a correspondence between topics and variables, and the coding matrix $W$ weights how much each topic is represented in the samples. While this decomposition applies directly to data when each sample is structured as a vector, it fails to accommodate data with tensor-like structure without reshaping. For example, flattening an image into a vector removes the spatial structure contained implicitly within the tensor. One approach to address this problem is the Non-negative CANDECOMP/PARAFAC (CP) tensor Decomposition (NCPD), which extends NMF to handle tensor data by approximating a mode-$n$ data tensor $\mathcal{A} \in \R^{\prod_{k=1}^n d_k}$ by rank-$r$ components \cite{carroll1970analysis, harshman1970foundations}. In particular, NCPD solves the minimization problem

\begin{equation}
    \argmin_{u^{k}_{j}}\|\mathcal{A} - \sum_{j=1}^r \otimes_{k=1}^n u^{k}_{j} \|_F^2
\label{eqn:ncpdloss}
\end{equation}
where $u^{k}_{j}\in \R^{d_k}$ denotes the $k$th vector used to form the $j$th rank-one outer product. In this work, we use \textit{rank} to denote the CP-rank \cite{kolda2009tensor}. 

In some settings it is desirable to account for data obtained from different sources. For example, data from hospitals across different states reflect separate populations, or experiments run on different equipment may have varying error distributions. In these cases we may wish to use \textit{stratified} methods, which can learn both strata-dependent information as well as shared information across different sources \cite{hansen1953sample}. The recent Stratified-NMF method extends NMF to consider stratified data that share global topics but differ by positive strata-dependent shifts, and demonstrates improved performance across data modalities \cite{chapman2023stratified}.  

Given $s$ strata and data matrices for each stratum $A(i),\;1 \leq i \leq s$, Stratified-NMF minimizes the objective
$$\argmin_{v(i), W(i), H} \sum_{i = 1}^{s}\|A(i) - \mathbf{1}v(i)^T - W(i)H\|_F^2,$$ 
such that
$$v(i)^T\geq 0,\; W(i)\geq 0,\; H\geq 0,\; \forall 1\leq i \leq s,$$
where $\mathbf{1}_m\in \R^m$ is the vector of all ones, $v(i)$ is the strata-dependent shift for each stratum, $W(i)$ is the coding matrix for each stratum, and $H$ is the global topics matrix. In this formulation, topics are learned across strata globally, while also allowing for the model to learn characteristics unique to each strata in the vectors $v(i)$ (strata features). 

In this paper, we create a Stratified-NTF method that extends Stratified-NMF to tensor data in order to leverage both geometric information contained in tensor data as well as stratification to handle heterogeneous datasets. We derive multiplicative updates to efficiently optimize the objective as is common of NMF and its variants. In experiments, we demonstrate the effectiveness and interpretability of the method and compare it to existing algorithms. Finally, we derive multiplicative updates for Stratified-NTF with TV-regularization and evaluate it on noisy image data. The code for this paper is made publicly available\footnote{Code can be found at https://github.com/alexandersietsema/Stratified-NTF}.

\section{Proposed Method}

Given a non-negative set of data tensors $\mathcal{A}(i)\in \R^{d_1(i)\times \prod_{k=2}^n d_k}$, 
we propose the following Stratified-NTF objective:
\begin{equation}
    \label{eqn:sntf_tensor}
    \argmin_{\mathcal{V}(i), w(i)_j, \mathcal{H}_j}
    \sum_{i=1}^s \| \mathcal{A}(i) - \mathbf{1}_{d_1(i)} \otimes \mathcal{V}(i) - \sum_{j=1}^r w(i)_j\otimes \mathcal{H}_{j}\|^2_F,
\end{equation}
where \begin{equation*}
    \mathcal{V}(i) = \sum_{j=1}^{r'(i)} 
    \mathcal{V}(i)_j,\ \ \ \mathcal{V}(i)_j = \otimes_{k=2}^{n} \; v(i)_{j}^k,\ \ \ \mathcal{H}_j= \otimes_{k=2}^{n} \; h_{j}^k,
\end{equation*}
such that \begin{equation*}
    v(i)_{j}^k \in \R^{d_k},\: w(i)_j \in \R^{d_1(i)},\: h_j^k \in \R^{d_k},\ \forall i, j, k.
\end{equation*}

We define $\mathcal{V}(i)_j$ to be the rank-one strata tensors that we sum over to get rank $r'(i)$ strata features $\mathcal{V}(i)$. Note that $\mathcal{V}(i)$ produces strata tensors analogous to the strata features $v(i)$ obtained in Stratified-NMF which learn information common to each sample in the stratum.

We additionally extend the method to include higher-rank decomposition of the strata-dependent shifts. This is necessary in the tensor setting since rank-one feature information may underfit the data \cite{shashua2005non}. For example, when modeling image data, a single outer product can only generate axis-aligned pixel clusters in the image. By increasing the rank, the strata features can learn sums of such clusters to represent more complex strata specific information. 

We derive multiplicative updates to efficiently optimize the objective. Let 

\begin{equation}
    \label{eqn:B_tensor_approx}
    \mathcal{B}(i) = \mathbf{1}_{d_1(i)} \otimes \mathcal{V}(i)+ \sum_{j=1}^r w(i)_j\otimes \mathcal{H}_{j}
\end{equation}
be the Stratified-NTF approximation of $\mathcal{A}(i)$. Then the multiplicative updates are
\begin{equation}
    \label{eqn:update_v}
    v(i)^{\tau}_{l k} \leftarrow v(i)^{\tau}_{l k} \frac{\sum_j\sum_{\alpha\in S^\tau} \mathcal{V}(i)_{l,\alpha} \mathcal{A}(i)_{j, \alpha+k e_\tau}}{\sum_j\sum_{\alpha\in S^\tau} \mathcal{V}(i)_{l,\alpha} \mathcal{B}(i)_{j, \alpha+k e_\tau}}
\end{equation}
\begin{equation}
    \label{eqn:update_w}
    w(i)_{l k} \leftarrow w(i)_{l k} \frac{\sum_{\alpha\in S} \mathcal{H}_{l,\alpha} \mathcal{A}(i)_{k, \alpha}}{\sum_{\alpha\in S} \mathcal{H}_{l,\alpha} \mathcal{B}(i)_{k, \alpha}}
\end{equation}
\begin{equation}
    \label{eqn:update_h}
    h_{l k}^\tau \leftarrow h_{l k}^\tau \frac{\sum_{ij}\sum_{\alpha\in S^\tau} (w(i)_l\otimes \mathcal{H}_l)_{j, \alpha}\mathcal{A}(i)_{j, \alpha+k e_\tau}}{\sum_{ij}\sum_{\alpha \in S^\tau} (w(i)_l\otimes \mathcal{H}_l)_{j,\alpha} \mathcal{B}(i)_{j, \alpha+k e_\tau}}
\end{equation}
where $S$ is the set of length $n-1$ multi-indices, and $S^{\tau}$ is the set of length $n-1$ multi-indices with the $\tau$th entry $0$. We use the convention that the zeroth index of a vector is defined to be $1$ for ease of notation and we use $e_j$ to denote the $j$th standard basis vector. These updates are written entry-wise for each vector and computed for all $l,k,\tau$ indexed as in \eqref{eqn:sntf_tensor}. The $\mathcal{B}(i)$ tensor is computed based on the existing values as in \eqref{eqn:B_tensor_approx} in each update step.

\subsection{Regularization}

In this section, we develop a Total Variation (TV) regularized version of Stratified-NTF for image denoising \cite{rudin1994total}. TV regularization is widely utilized in image processing and has been effectively applied in tensor decompositions for promoting smoothness in applications such as hyperspectral unmixing \cite{hyperspec}. Unlike NMF methods, which require flattening images into a single mode, tensor methods retain spatial information which we can directly apply regularization to. We apply regularization to the global topic parameters $\mathcal{H}$ of form
\begin{equation}
\label{eqn:sntf_tensor_reg}
    ||h^{\tau}_l||_{TV} = \sum_k |h_{l, k+1}^\tau - h_{l, k}^\tau|.
\end{equation}
For application to 3-mode image data, we regularize separately over $h^2$ and $h^3$, the spatial modes of $\mathcal{H}$,  as
\begin{equation}
\begin{split}
    \argmin_{\mathcal{V}(i), w(i)_j, \mathcal{H}_j}
    \sum_{i=1}^s \| \mathcal{A}(i) - \mathbf{1}_{d_1(i)} &\otimes \mathcal{V}(i) - \sum_{j=1}^r w(i)_j\otimes \mathcal{H}_{j}\|^2_F \\
    &+\lambda \sum_l \left(||h^2_l||_{TV}+||h^3_l||_{TV}\right).
\end{split}
\end{equation}
The gradient of the TV term for fixed $\tau$ is 
\begin{equation}
    \nabla ||h_l^\tau||_{TV} = \begin{bmatrix}
           -\text{sign}(h^{\tau}_{l,1} - h^{\tau}_{l,0})\\
           - \text{sign}(h^{\tau}_{l,2} - h^{\tau}_{l,1}) + \text{sign}(h^{\tau}_{l,1} - h^{\tau}_{l,0}) \\
           - \text{sign}(h^{\tau}_{l,3} - h^{\tau}_{l,2}) + \text{sign}(h^{\tau}_{l,2} - h^{\tau}_{l,1}) \\
           \vdots \\
           - \text{sign}(h^{\tau}_{l,k+1} - h^{\tau}_{l,k}) + \text{sign}(h^{\tau}_{l,k} - h^{\tau}_{l,k-1}) \\
           \vdots \\
            \text{sign}(h^{\tau}_{l,d_\tau} - h^{\tau}_{l,d_\tau-1}) \\
         \end{bmatrix}.
\end{equation}
We define $\nabla^+||h_l^\tau||_{TV}$ and $\nabla^-||h_l^\tau||_{TV}$ to be the vector of positive and negative entries of the gradient, respectively, and zeros elsewhere as
\begin{align}
\nabla^+||h_l^\tau||_{TV}&= \max\left\{0, \nabla||h_l^\tau||_{TV}\right\}\\
\nabla^-||h_l^\tau||_{TV}&= - \min\left\{ \nabla||h_l^\tau||_{TV},0\right\},
\end{align} 
where $\max$ and $\min$ are evaluated entry-wise. This results in the regularized multiplicative update equation

\begin{equation}
    \label{eqn:update_h_reg}
    \!h_{l k}^\tau \!\leftarrow\! h_{l k}^\tau \!\frac{\sum_{ij}\!\sum_{\alpha\in S^\tau} \!(w(i)_l\!\otimes\! \mathcal{H}_l)_{j, \alpha}\mathcal{A}(i)_{j, \alpha+k e_\tau}\!+\!\lambda\!\nabla^-||h_l^\tau||_{TV}}{\sum_{ij}\!\sum_{\alpha \in S^\tau} \!(w(i)_l\!\otimes\! \mathcal{H}_l)_{j,\alpha} \mathcal{B}(i)_{j, \alpha+k e_\tau}\!+\!\lambda\!\nabla^+||h_l^\tau||_{TV}}.
\end{equation}

We normalize each $h_l^\tau$ after each update to stabilize the method. This is a standard technique in $\ell_1$-norm regularization for NMF methods \cite{eggert2004sparse}. The full algorithm is shown in Algorithm~\ref{alg:stratifiedntf}.

\begin{algorithm}
\caption{Stratified-NTF Multiplicative Update Algorithm}\label{alg:stratifiedntf}
\begin{algorithmic}[1]
\State{\textbf{Input:} Data $\mathcal{A}(i)\in \R^{d_1(i)\times \prod_{k=2}^n d_k}$ for $1 \leq i \leq s$, decomposition ranks $r, r'(i)$, number of iterations $N$, number of strata updates per iteration $M$, regularization strength $\lambda$}
\State{Initialize $v(i)_{j'}^k\in \R^{d_k}, w(i)_j\in \R^{d_1(i)}, h_j^k\in \R^{d_k}$ for \newline $1 \leq j \leq r,\; 1 \leq j' \leq r'(i)$}
\For{$N$ iterations}
\For{$M$ iterations }
\For{$\tau$ modes}

\State{Update $v(i)^\tau\ \forall i$ in each entry according to \eqref{eqn:update_v}}
\EndFor{}
\EndFor{}
\State{Update $w(i)_j\ \forall i,j$ in each entry and rank according to \eqref{eqn:update_w}}

\For{$\tau$ modes}
    \If{regularization}
        \State {Update $h^\tau_j\ \forall j$ in each entry according to \eqref{eqn:update_h_reg}}
        \State Normalize $h^\tau_j\ \forall j$

    \Else
        \State Update $h^\tau_j\ \forall j$ in each entry according to \eqref{eqn:update_h}
    \EndIf
\EndFor
\EndFor{}
\end{algorithmic}
\end{algorithm}

\section{Experimental Results}

In this section, we demonstrate the effectiveness and interpretability of Stratified-NTF applied to text and image data. When computing the updates \eqref{eqn:update_v}, \eqref{eqn:update_w}, \eqref{eqn:update_h}, we clip each numerator and denominator to machine precision tolerance to avoid zero values for numerical stability \cite{tensorly}. All parameters are randomly initialized from independent and identically distributed $[0,1]$ uniform distributions. We compute $\mathcal{V}(i)$ updates twice in each iteration, corresponding to the choice $M=2$ in Line 4 of Algorithm \ref{alg:stratifiedntf}.

\subsection{20 Newsgroups}

We first apply Stratified-NTF to the 20 newsgroups dataset \cite{20newsgroups}. 
The dataset contains document data from 20 newsgroups of various super-categories and sub-categories as summarized in Table \ref{tab:topics}. We preprocess the data by removing headers, footers, and quotes, and applying TF-IDF vectorization for the top 5000 words, ignoring words which appear in over 95\% of the documents. We construct our dataset with strata consisting of the five super-categories, and for each stratum we construct the data tensor with axes specified by
$$\text{sub-category}\times \text{document} \times \text{word}.$$

The learned parameters of the Stratified-NTF method should thus represent the correspondence between each learned topic and each corresponding mode. In particular, $w(i)_j$ learns the correspondence between the $j$th topic and each sub-category for a super-category $i$, $h^2_j$ learns the correspondence between the $j$th topic and each document, and $h^3_j$ learns the correspondence between the $j$th topic and each word. This three-mode representation allows us to characterize topics in finer detail than is possible with existing matrix methods, which have only two modes in which to learn topic information. As the factorization is non-negative, these parameters yield a naturally interpretable linear representation of the data tensor as a product of the sub-category, document, and word correspondences.

In order for the document dimension to align between tensors of different strata, 
we choose 100 documents for each sub-category. We apply Stratified-NTF to this dataset with strata rank 3 and topic rank 5 for 11 iterations. Because of the natural sparsity of text data, we observe almost immediate convergence.

\begin{table}[]
\centering
\begin{tabular}{llll} \toprule
Super-category & Sub-categories \\ 
\midrule
    \multirow[c]{2}{*}{Computers} &  Graphics & MS Windows& PC Hardware \\
            & Mac Hardware& X Windows &\\
    \cline{2-4}
    \multirow[c]{2}{*}{Recreation} & Autos & Motorcycles& Baseball\\
    & Hockey & & \\
    \cline{2-4}
    \multirow[c]{2}{*}{Science} & Crypto & Electronics & Medicine  \\ 
    & Space &  & \\
    \cline{2-4}
    Politics & Guns & Middle East & Miscellaneous\\ 
    \cline{2-4}
    Religion & Atheism & Christianity & Miscellaneous\\
    \bottomrule
\end{tabular}
\caption{Super-categories and sub-categories for the 20 Newsgroups dataset.}
\label{tab:topics}
\end{table}

In Table \ref{tab:topics}, we identify the top three highest correspondence words for each strata feature, that is, the words corresponding to the three highest values in each $v(i)_j^3$. We see that the strata features capture relevant topic information for each stratum. For example, the ``Recreation'' stratum contains features corresponding to the sports-related ``Hockey'' and ``Baseball'' sub-categories (the feature ``Players'', ``Year'', ``Flyers'') as well as to the ``Motorcycles'' sub-category (the feature ``Bike'', ``Miles'', ``Game''). Similarly, the ``science'' stratum contains a feature related to the ``Crypto'' sub-category (the feature ``Key'', ``Private'', ``Keys''). 

Similarly, we can identify the highest correspondence documents for each strata feature by finding the largest values in the document mode for $v(i)_j^2$. We see that such documents are indeed topical: for example, for the hockey-related feature (``Players'', ``Year'', ``Flyers''), the document corresponding to the largest value of $v(2)_1^2$ in the ``Hockey'' mode discusses a potential player trade between the Philadelphia Flyers and Quebec Nordiques in the National Hockey League. While this requires human interpretation of the strata features to identify the correct sub-category mode in which to find the related document, this demonstrates that the learned strata features also reflect the content of individual documents. Overall, we observe that for this dataset, the method is able to identify interpretable strata features that are both relevant to the super-category of each stratum, as well as relevant to the sub-categories contained in each stratum's data tensor.

\begin{table}[]
\centering
\begin{tabular}{clll}\toprule
    Super-category & 1 & 2 & 3\\
    \midrule
    \multirow[c]{3}{*}{Computers} & Thanks & Use & Mail\\
    
    & Files & Apps & Windows\\
    & Window & Tiff & Problem\\
     \cline{2-4}
    \multirow[c]{3}{*}{Recreation} & Players & Year & Flyers\\
     & Bike & Miles & Game\\
     & Car & Think & Ya\\
     \cline{2-4}
     \multirow[c]{3}{*}{Science}& Key & Private & Keys\\
     & Clipper & Chip & Encryption\\
     & Space & Electronics & Amp\\
     \cline{2-4}
     \multirow[c]{3}{*}{Politics}
     & Armenians & Armenian & Firearms\\
     & Turkish & Gun & Government\\
     & Armenian & Armenians & Gay\\  
    \cline{2-4}
    \multirow[c]{3}{*}{Religion}
     & Sin & People & Christian\\
     & God & Jesus & Father\\
     & Objective & Morality & Moral\\ 
     \bottomrule
\end{tabular}
\caption{The top three most relevant words learned for each strata feature. The three rows within a super-category $i$ correspond to each of $v(i)^3_1, v(i)^3_2, v(i)^3_3$.}
\label{tab:20newsgroups}
\end{table}

\subsection{Olivetti Faces}

In this section, we compare the performance of NTF, Stratified-NMF, and Stratified-NTF on the Olivetti Faces Dataset constructed by AT\&T Laboratories Cambridge \cite{samaria1994parameterisation} as provided by the scikit-learn package \cite{scikit-learn}. The Olivetti Faces Dataset consists of 10 images of each of 40 individuals taken from multiple angles all of dimension $64 \times 64$. To create our stratified dataset, we stratify over each individual to create 40 data tensors where each stratum is of dimension $10 \times 64 \times 64$.

As a baseline comparison, we use a standard NTF implementation from Tensorly \cite{tensorly} with the loss updated to provide loss values matching \eqref{eqn:ncpdloss}. Since NTF is not a stratified method, we input the data tensor concatenated across strata of shape $400 \times 64 \times 64$. For Stratified-NMF, we flatten the data tensor across the last two dimensions to create data matrices for each stratum of shape $10 \times 4096$. In order to make fair comparisons across methods, we pick hyperparameters for each method to yield as close to the same total number of learnable parameters, as shown in Table~\ref{tab:hparam}. Note that as Stratified-NMF uses flattened data, each rank is an entire image, so the closest possible comparison uses a strata feature rank of 1 (fixed in Stratified-NMF) and a topic rank of 1 for a total of $\sim$168K parameters. 

\begin{table}[]
    \centering
    \begin{tabular}{rccc} \toprule
        & Stratified-NTF & NTF & Stratified-NMF\\ \midrule
        Topic rank & 40 & 186 & 1\\
        Strata rank & 15 & - & 1\\
        Total params & $\sim$97K & $\sim$98K & $\sim$168K \\ 
        \bottomrule
    \end{tabular}
    \caption{Learnable parameters for Stratified-NTF, NTF, and Stratified-NMF on the faces experiment.}
    \label{tab:hparam}
\end{table}

\begin{figure}
    \centering
    \includegraphics[width=0.7\linewidth,trim={0cm 0cm 0cm 1cm}, clip]{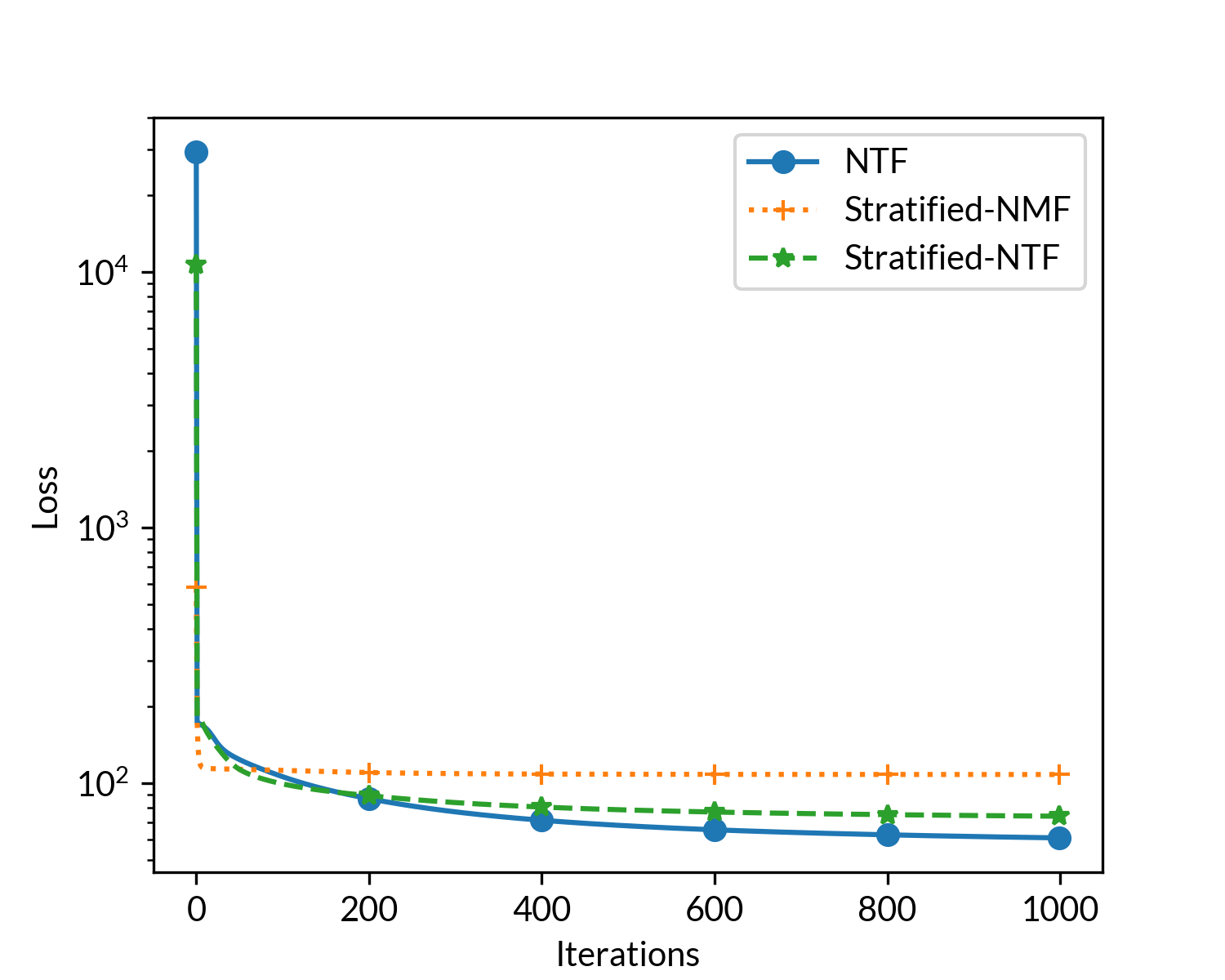}
    \caption{Loss plots for NTF, Stratified-NMF, and Stratified-NTF run for 1000 iterations on the Olivetti faces dataset. The rapid convergence shows that multiplicative updates are able to efficiently optimize the objective, matching existing methods.}
    \label{fig:faceslossplot}
\end{figure}

Figure~\ref{fig:faceslossplot} compares the convergence of each method over 1000 iterations. The multiplicative update method for Stratified-NTF described in Algorithm $\eqref{alg:stratifiedntf}$ converges on the Olivetti dataset with a final loss of $74.115$. We see nearly immediate convergence, which highlights the efficiency of the multiplicative updates. This outperforms Stratified-NMF, which obtains a final loss of $107.921$ on this dataset. We also see that NTF modestly outperforms Stratified-NTF. This is not unexpected: though the total number of parameters is similar for each method, for a particular image, Stratified-NTF has access to only $65$ total topics (the global topics and one set of strata features). On the other hand, NTF has access to all $186$ topics, so NTF has more flexibility for a given image. However, these performance gains are offset by the added interpretability of learning both global and local information as provided by stratified methods. 

\begin{figure}
\centering
  \begin{subfigure}[b]{0.9\columnwidth}
  \centering
    \includegraphics[width=0.4\linewidth]
    {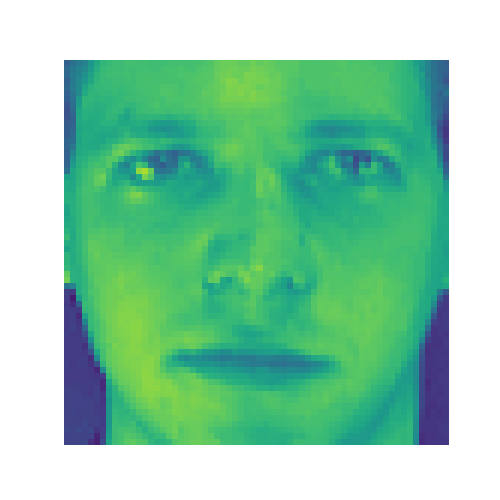}
    \hspace{-1em}
    \includegraphics[width=0.4\linewidth]{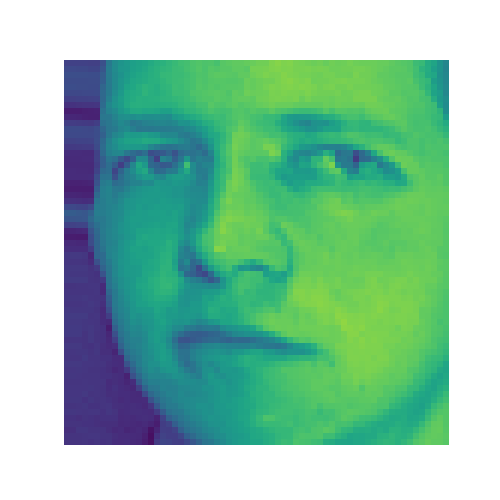}
    \vspace{-1em}
    \caption{Ground truth}
    \label{fig:comparisongroundtruth}
  \end{subfigure}
  \vskip \baselineskip \vspace{-1.5em}
  \begin{subfigure}[b]{0.9\columnwidth}
  \centering
    \includegraphics[width=0.4\linewidth]
    {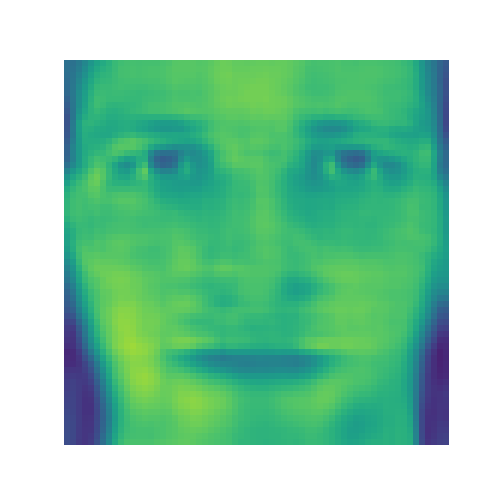}
    \hspace{-1em}
    \includegraphics[width=0.4\linewidth]{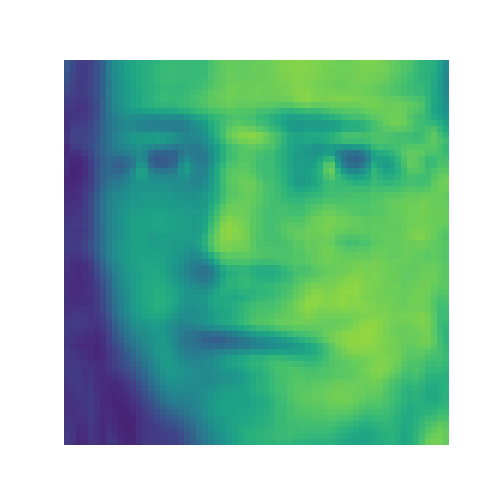}
    \vspace{-1em}
    \caption{NTF}
    \label{fig:comparisonntf}
  \end{subfigure}
  \vskip \baselineskip \vspace{-1.5em}
  \begin{subfigure}[b]{0.9\columnwidth}
  \centering
    \includegraphics[width=0.4\linewidth]
    {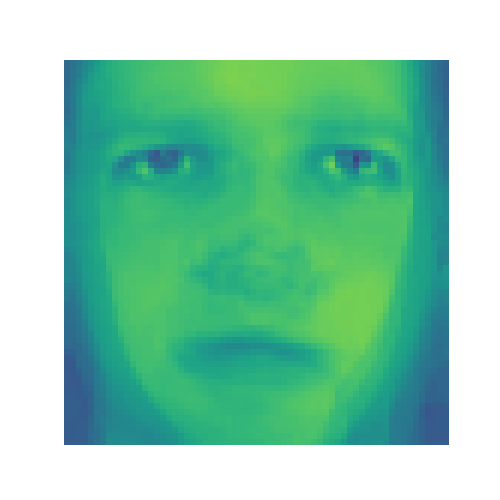}
    \hspace{-1em}
    \includegraphics[width=0.4\linewidth]{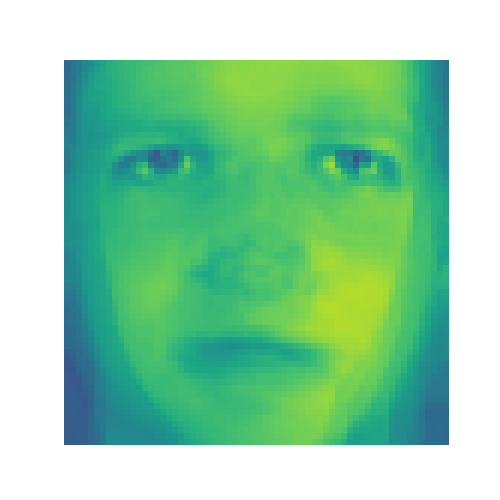}
    \vspace{-1em}
    \caption{Stratified-NMF}
    \label{fig:comparisonsnmf}
  \end{subfigure}
  \vskip \baselineskip \vspace{-1.5em}
  \begin{subfigure}[b]{0.9\columnwidth}
  \centering
    \includegraphics[width=0.4\linewidth]
    {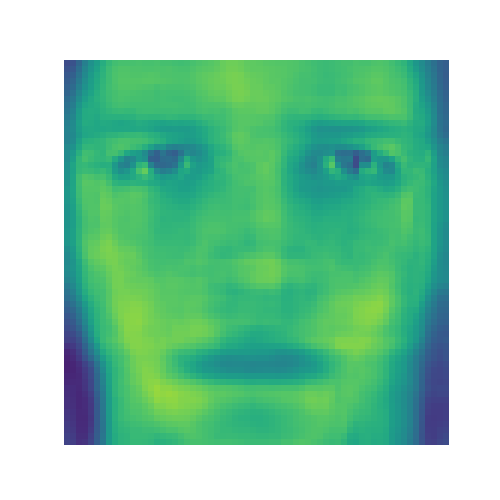}
    \hspace{-1em}
    \includegraphics[width=0.4\linewidth]{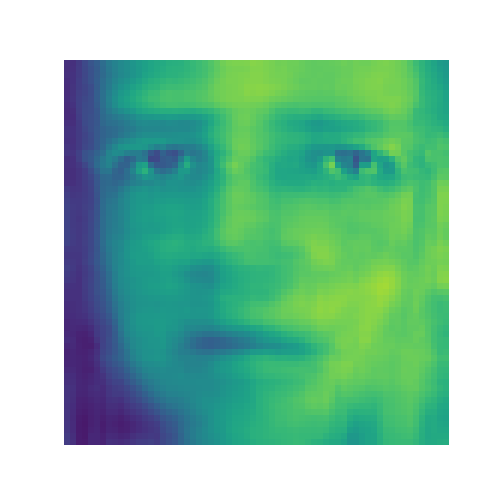}
    \vspace{-1em}
    \caption{Stratified-NTF}
    \label{fig:comparisonsntf}
  \end{subfigure}
  \caption{Reconstruction comparison of NTF, Stratified-NMF, and Stratified-NTF on the first (left) and fourth (right) images from the first stratum. The NTF and Stratified-NTF reconstructions appear qualitatively similar, while the Stratified-NMF reconstructions fail to capture the different viewing angle between the two images due to differences in parameterization.}
  \label{fig:facescomparison}
\end{figure}
\

Figure~\ref{fig:facescomparison} shows a comparison of the resulting reconstructions from each method for the first pair of images of the first individual with different viewing angles. Qualitatively, we see that NTF and Stratified-NTF perform similarly, while the Stratified-NMF reconstruction is far less detailed and displays color banding around the edges of the face.
Additionally, reconstructed images for Stratified-NMF appear front-facing while the tensor methods are able to capture the change in head angle. This highlights the added expressiveness of Stratified-NTF compared to Stratified-NMF, even while using only just over half the same number of parameters. This indicates that the tensor methods may be more memory-efficient for this dataset.

\subsection{Noisy, watermarked MNIST}

To evaluate the performance of Stratified-NTF with TV-regularization, we conduct an experiment on a noisy, watermarked version of the MNIST dataset \cite{lecun2010mnist}. 
The MNIST dataset consists of images of handwritten digits from the Torchvision library \cite{marcel2010torchvision}, which we rescale to floating-point values between 0 and 1. 
To construct our experiment data, we create two strata: the first consists of $100$ images of 1's and $100$ images of 2's, all with a text watermark overlay of the word ``ONE'' positioned at the top of the image. The second stratum consists of $100$ images of 2's and $100$ images of 3's, all with a text watermark overlay of the word ``TWO'' at the bottom of the image. We add salt-and-pepper noise to all images by randomly setting pixels of each image to 0 and 1, both with probability $0.15$. We note that TV regularization is a standard technique for removing salt-and-pepper noise \cite{rodriguez2013total, nikolova2004variational}. We run TV-regularized Stratified NTF on this dataset across regularization parameters of $\lambda = 0.0, 5.0, 10.0$, each for $100$ iterations with strata feature rank $100$ and topic rank $100$. 

As the text watermarks are present uniformly in each image of each stratum, we expect the strata features to identify and separate out the watermarks, effectively performing watermark removal on the dataset. As the strata differ in the contained digits as well as the watermarks, Stratified-NTF must capture both the fine detail of the text as well as the larger forms of the digits. By regularizing only the topic parameters, we expect to denoise the digits while not affecting text learned by the strata features. Unlike the topic parameters, the strata features are added uniformly to each image in each stratum, so they cannot overfit to the image-specific noise patterns. Therefore regularization should only be necessary in the $\mathcal{H}_j$ tensors to achieve denoising in all of the learned parameters.

\begin{figure}
\centering
    \begin{subfigure}[b]{0.9\columnwidth}
  \centering
    \includegraphics[width=0.35\linewidth]
    {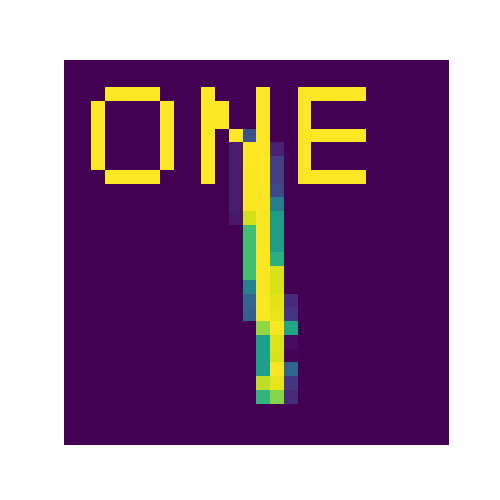}
    \hspace{-1em}
    \includegraphics[width=0.35\linewidth]{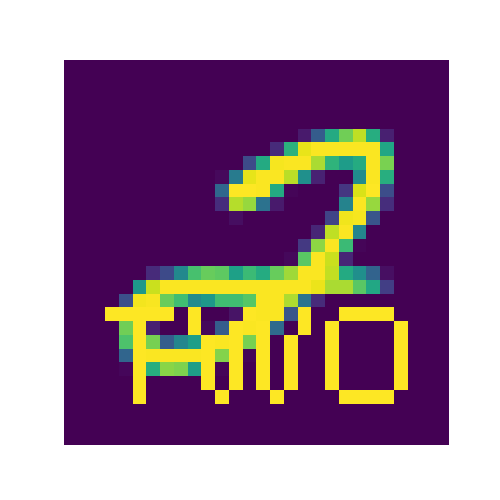}
    \vspace{-1em}
    \caption{Watermarked Data}
    \label{fig:reggroundtruth}
  \end{subfigure}
  \begin{subfigure}[b]{0.9\columnwidth}
  \centering
    \includegraphics[width=0.35\linewidth]
    {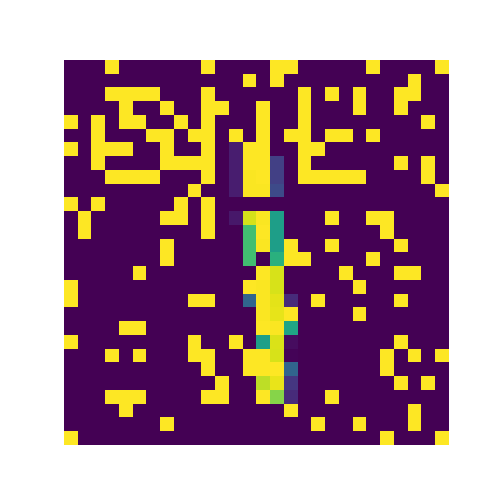}
    \hspace{-1em}
    \includegraphics[width=0.35\linewidth]{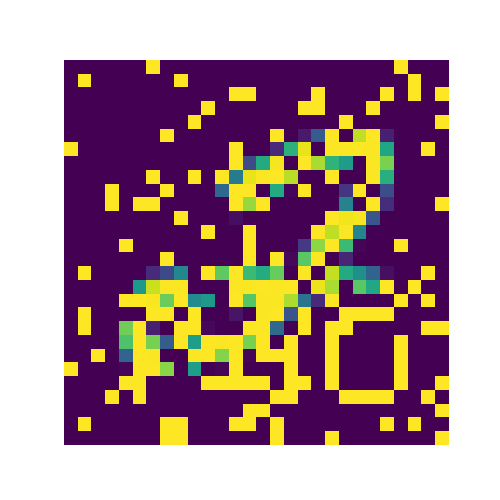}
    \vspace{-1em}
    \caption{Noisy Watermarked Data}
    \label{fig:reggroundtruthnoisy}
  \end{subfigure}
  \vskip \baselineskip \vspace{-1.3em}
  \begin{subfigure}[b]{0.9\columnwidth}
  \centering
    \includegraphics[width=0.35\linewidth]
    {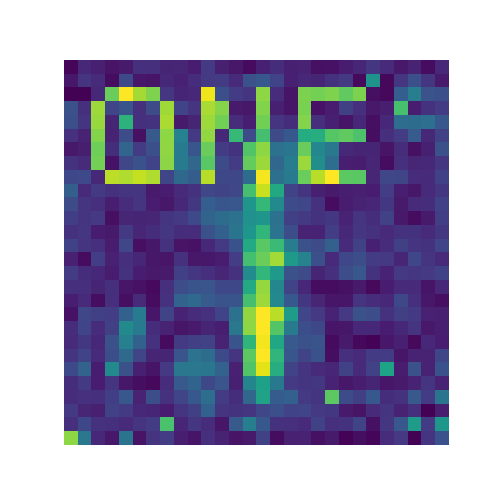}
    \hspace{-1em}
    \includegraphics[width=0.35\linewidth]{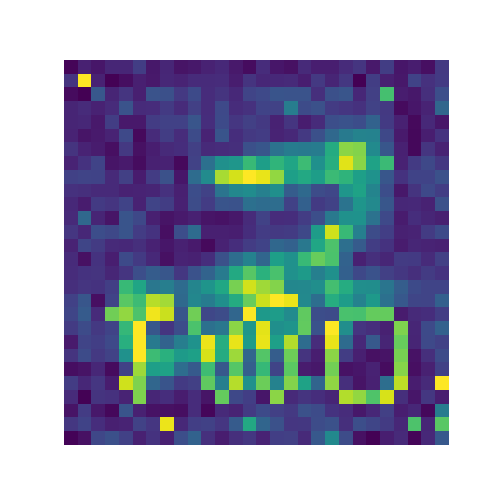}
    \vspace{-1em}
    \caption{$\lambda = 0.0$}
    \label{fig:0reconstruction}
  \end{subfigure}
  \vskip \baselineskip \vspace{-1.5em}
  \begin{subfigure}[b]{0.9\columnwidth}
  \centering
    \includegraphics[width=0.35\linewidth]
    {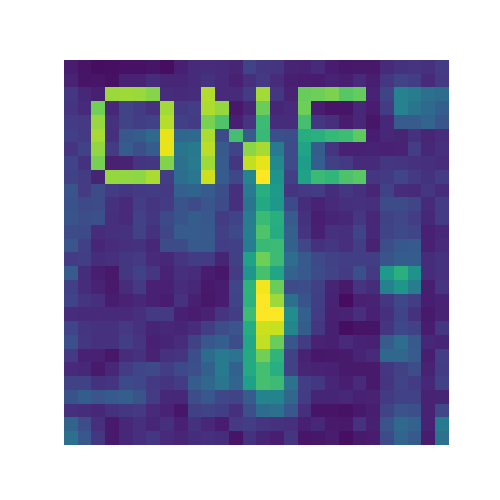}
    \hspace{-1em}
    \includegraphics[width=0.35\linewidth]{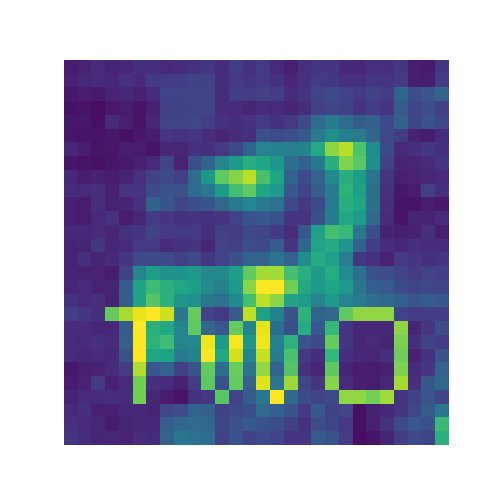}
    \vspace{-1em}
    \caption{$\lambda = 5.0$}
    \label{fig:3reconstruction}
  \end{subfigure}
  \vskip \baselineskip \vspace{-1.5em}
  \begin{subfigure}[b]{0.9\columnwidth}
  \centering
    \includegraphics[width=0.35\linewidth]
    {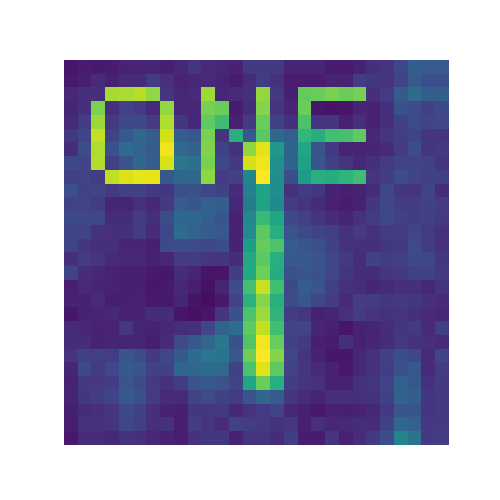}
    \hspace{-1em}
    \includegraphics[width=0.35\linewidth]{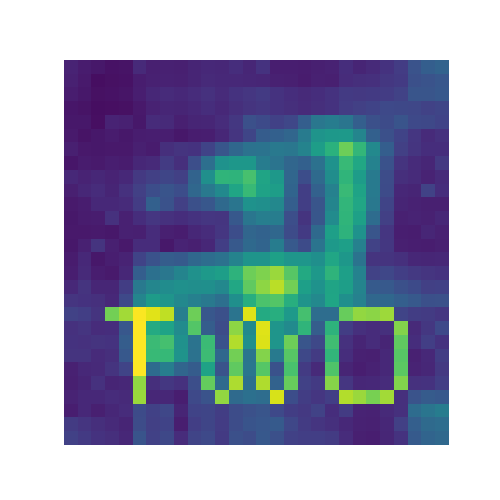}
    \vspace{-1em}
    \caption{$\lambda = 10.0$}
    \label{fig:10reconstruction}
  \end{subfigure}
  \caption{Reconstruction comparison of TV-regularized Stratified-NTF on noisy watermarked data across different regularization parameters $\lambda$. We observe that stronger regularization decreases noise in the resulting reconstructions.}
  \label{fig:regcomparisonsntf}
\end{figure}

In Fig. \ref{fig:regcomparisonsntf}, we compare the reconstructions of the first image in each stratum across noise levels. We observe that increasing the regularization parameter significantly reduces the amount of noise present in the reconstructions, while also preserving the text watermarks. The digit 1 in the first stratum images (left) is significantly clearer with higher regularization parameters, while the digit 2 in the second stratum (right) is quite blurry. We suspect that this is due to the shape of the digits interacting with the total variation of the topics: the digit 1 is nearly convex, which has a naturally lower total variation than the top and loop of the digit 2, so the model may have an easier time reconstructing the 1 in the presence of the regularization, especially given the high level of corruption in the data. 

Overall, while this formulation of TV regularization does not lead to dramatic improvements in reconstruction quality, we are still able to see the desired denoising effects. We note that this is a simple example to demonstrate the flexibility of mode-specific regularization compared to matrix methods like Stratified-NMF; there are a number of spatial and temporal regularization techniques for tensor factorizations which may be introduced \cite{anderson2017supervised}, as well as classical regularization approaches for NMF which may be extended to the tensor case \cite{salehani2017smooth}.

\section{Discussion}

The extension of NTF and Stratified-NMF to Stratified-NTF allows practitioners to combine the benefits of both methods. Practitioners can now leverage geometric priors in tensor data through an explicit tensor representation and with regularization, while also handling heterogeneous data. This work also opens the door to some important open questions regarding 

\begin{itemize}
    \item Principled understanding of hyperparameter choices
    \item Investigation of additional regularization methods
    \item Regularization interacting between strata features and global features
    \item Proofs of convergence of the multiplicative updates \cite{shashua2005non, chapman2023stratified, lin2007convergence}.
\end{itemize}

\section{Conclusion}
In this work, we extend the Stratified-NMF method into a stratified non-negative tensor decomposition. Stratified-NTF can learn both global and local information from multi-modal data from different sources. We develop multiplicative update rules for the method as well as for a total-variation regularized variant. We demonstrate the effectiveness of the standard method on text and image data, and demonstrate the regularized method's effectiveness for noise reduction on noisy, watermarked image data. 

\bibliographystyle{IEEEtran}
\bibliography{IEEEabrv,ref}

\begin{thebibliography}{10}
\providecommand{\url}[1]{#1}
\csname url@samestyle\endcsname
\providecommand{\newblock}{\relax}
\providecommand{\bibinfo}[2]{#2}
\providecommand{\BIBentrySTDinterwordspacing}{\spaceskip=0pt\relax}
\providecommand{\BIBentryALTinterwordstretchfactor}{4}
\providecommand{\BIBentryALTinterwordspacing}{\spaceskip=\fontdimen2\font plus
\BIBentryALTinterwordstretchfactor\fontdimen3\font minus \fontdimen4\font\relax}
\providecommand{\BIBforeignlanguage}[2]{{%
\expandafter\ifx\csname l@#1\endcsname\relax
\typeout{** WARNING: IEEEtran.bst: No hyphenation pattern has been}%
\typeout{** loaded for the language `#1'. Using the pattern for}%
\typeout{** the default language instead.}%
\else
\language=\csname l@#1\endcsname
\fi
#2}}
\providecommand{\BIBdecl}{\relax}
\BIBdecl

\bibitem{shashua2005non}
A.~Shashua and T.~Hazan, ``Non-negative tensor factorization with applications to statistics and computer vision,'' in \emph{Proc. 22nd Int. Conf. Mach. Learn.}, 2005, pp. 792--799.

\bibitem{lee2000algorithms}
D.~Lee and H.~S. Seung, ``Algorithms for non-negative matrix factorization,'' \emph{Adv. Neural Inf. Process. Syst.}, vol.~13, 2000.

\bibitem{lee1999learning}
D.~D. Lee and H.~S. Seung, ``Learning the parts of objects by non-negative matrix factorization,'' \emph{Nature}, vol. 401, no. 6755, pp. 788--791, 1999.

\bibitem{carroll1970analysis}
J.~D. Carroll and J.-J. Chang, ``Analysis of individual differences in multidimensional scaling via an n-way generalization of “{E}ckart-{Y}oung” decomposition,'' \emph{Psychometrika}, vol.~35, no.~3, pp. 283--319, 1970.

\bibitem{harshman1970foundations}
R.~A. Harshman \emph{et~al.}, ``Foundations of the {PARAFAC} procedure: Models and conditions for an “explanatory” multi-modal factor analysis,'' \emph{UCLA Work. Pap. Phonetics}, vol.~16, no.~1, p.~84, 1970.

\bibitem{kolda2009tensor}
T.~G. Kolda and B.~W. Bader, ``Tensor decompositions and applications,'' \emph{SIAM Rev.}, vol.~51, no.~3, pp. 455--500, 2009.

\bibitem{hansen1953sample}
M.~H. Hansen, W.~N. Hurwitz, and W.~G. Madow, \emph{Sample survey methods and theory. {V}ol. {I}. Methods and applications.}\hskip 1em plus 0.5em minus 0.4em\relax John Wiley, 1953.

\bibitem{chapman2023stratified}
J.~Chapman, Y.~Yaniv, and D.~Needell, ``Stratified-{NMF} for heterogeneous data,'' in \emph{Asilomar Conf. Signals Syst. Comp.}\hskip 1em plus 0.5em minus 0.4em\relax IEEE, 2023, pp. 614--618.

\bibitem{rudin1994total}
L.~I. Rudin and S.~Osher, ``Total variation based image restoration with free local constraints,'' in \emph{IEEE Image Proc.}, vol.~1.\hskip 1em plus 0.5em minus 0.4em\relax IEEE, 1994, pp. 31--35.

\bibitem{hyperspec}
F.~Xiong, Y.~Qian, J.~Zhou, and Y.~Y. Tang, ``Hyperspectral unmixing via total variation regularized nonnegative tensor factorization,'' \emph{IEEE Trans. Geosci. Remote Sens.}, vol.~57, no.~4, pp. 2341--2357, 2019.

\bibitem{eggert2004sparse}
J.~Eggert and E.~Korner, ``Sparse coding and {NMF},'' in \emph{Int. Jt. Conf. Neural Netw. (IEEE Cat. No. 04CH37541)}, vol.~4.\hskip 1em plus 0.5em minus 0.4em\relax IEEE, 2004, pp. 2529--2533.

\bibitem{tensorly}
\BIBentryALTinterwordspacing
J.~Kossaifi, Y.~Panagakis, A.~Anandkumar, and M.~Pantic, ``Tensor{L}y: Tensor learning in python,'' \emph{J. Mach. Learn. Res.}, vol.~20, no.~26, pp. 1--6, 2019. [Online]. Available: \url{http://jmlr.org/papers/v20/18-277.html}
\BIBentrySTDinterwordspacing

\bibitem{20newsgroups}
J.~Rennie and K.~Lang, ``Home page for 20 newsgroups dataset,'' 2008.

\bibitem{samaria1994parameterisation}
F.~S. Samaria and A.~C. Harter, ``Parameterisation of a stochastic model for human face identification,'' in \emph{Proc. Appl. Comput. Vis.}\hskip 1em plus 0.5em minus 0.4em\relax IEEE, 1994, pp. 138--142.

\bibitem{scikit-learn}
F.~Pedregosa, G.~Varoquaux, A.~Gramfort, V.~Michel, B.~Thirion, O.~Grisel, M.~Blondel, P.~Prettenhofer, R.~Weiss, V.~Dubourg, J.~Vanderplas, A.~Passos, D.~Cournapeau, M.~Brucher, M.~Perrot, and E.~Duchesnay, ``Scikit-learn: Machine learning in {P}ython,'' \emph{J. Mach. Learn. Res.}, vol.~12, pp. 2825--2830, 2011.

\bibitem{lecun2010mnist}
Y.~LeCun, C.~Cortes, and C.~Burges, ``{MNIST} handwritten digit database,'' \emph{ATT Labs. Available: http://yann.lecun.com/exdb/mnist}, vol.~2, 2010.

\bibitem{marcel2010torchvision}
S.~Marcel and Y.~Rodriguez, ``Torchvision the machine-vision package of torch,'' in \emph{Proc. ACM 18 Int. Conf. Multimedia}, 2010, pp. 1485--1488.

\bibitem{rodriguez2013total}
P.~Rodr{\'\i}guez, ``Total variation regularization algorithms for images corrupted with different noise models: a review,'' \emph{J. Electr. Comput. Eng.}, vol. 2013, no.~1, p. 217021, 2013.

\bibitem{nikolova2004variational}
M.~Nikolova, ``A variational approach to remove outliers and impulse noise,'' \emph{J. Math. Imaging Vision}, vol.~20, no.~1, pp. 99--120, 2004.

\bibitem{anderson2017supervised}
D.~Anderson, A.~Bapst, J.~Coon, A.~Pung, and M.~Kudenov, ``Supervised non-negative tensor factorization for automatic hyperspectral feature extraction and target discrimination,'' in \emph{Proc. SPIE}, vol. 10198.\hskip 1em plus 0.5em minus 0.4em\relax SPIE, 2017, pp. 263--275.

\bibitem{salehani2017smooth}
Y.~E. Salehani and S.~Gazor, ``Smooth and sparse regularization for {NMF} hyperspectral unmixing,'' \emph{IEEE Trans. Geosci. Remote Sens.}, vol.~10, no.~8, pp. 3677--3692, 2017.

\bibitem{lin2007convergence}
C.-J. Lin, ``On the convergence of multiplicative update algorithms for nonnegative matrix factorization,'' \emph{IEEE Trans. on Neural Netw.}, vol.~18, no.~6, pp. 1589--1596, 2007.

\end{thebibliography}

\end{document}